\documentclass[10pt,twocolumn,letterpaper]{article}
\newcommand{\vdi}{\textsc{{VDInstruct}}\xspace}

\usepackage[pagenumbers]{cvpr} 

\newcommand{\increasenoparent}[1]{\textcolor{ForestGreen}{+#1}}

\definecolor{cvprblue}{rgb}{0.21,0.49,0.74}
\usepackage[pagebackref,breaklinks,colorlinks,allcolors=cvprblue]{hyperref}
\usepackage{pifont}
\usepackage[capitalize]{cleveref}
\usepackage{xcolor}
\usepackage{tcolorbox}
\usepackage{colortbl} 
\usepackage{wrapfig} 
\usepackage{makecell}
\usepackage{enumitem}
\usepackage{amssymb}

\crefname{section}{Sec.}{Secs.}
\Crefname{section}{Section}{Sections}
\Crefname{table}{Table}{Tables}
\crefname{table}{Tab.}{Tabs.}

\def\paperID{*****} 
\def\confName{CVPR}
\def\confYear{2025}

\title{VDInstruct: Zero-Shot Key Information Extraction via Content-Aware Vision Tokenization}

\author{Son Nguyen\textsuperscript{\textdagger}, Giang Nguyen\textsuperscript{\ding{169}}, Hung Dao\textsuperscript{\textdagger},
Thao Do\textsuperscript{\textdagger}, Daeyoung Kim\textsuperscript{\textdagger}\\
KAIST, South Korea\textsuperscript{\textdagger}, Auburn University, US\textsuperscript{\ding{169}}\\
{\tt\small \{nguyendinhson,hicehehe,thaodo,kimd\}@kaist.ac.kr\textsuperscript{\textdagger}, nguyengiangbkhn@gmail.com\textsuperscript{\ding{169}}}
}

\begin{document}
\maketitle
\begin{abstract}
Key Information Extraction (KIE) underpins the understanding of visual documents (\eg, receipts and contracts) by extracting precise semantic content and accurately capturing spatial structure.
Yet existing multimodal large language models (MLLMs) often perform poorly on \emph{dense documents} and rely on vision tokenization approaches that scale with image size, leading to redundant computation and memory inefficiency.
To address these challenges, we introduce \vdi, an MLLM that separates spatial region detection from semantic feature extraction. 
Central to our model is a content‐aware tokenization strategy: rather than fragmenting the entire image uniformly, it generates tokens in proportion to document complexity, preserving critical structure while eliminating wasted tokens.
Leveraging a three‐stage training paradigm, our model achieves state‐of‐the‐art (SOTA) results on KIE benchmarks, matching or exceeding the accuracy of leading approaches while reducing the number of image tokens by roughly \textcolor{ForestGreen}{3.6$\times$}. 
In zero‐shot evaluations, \vdi surpasses strong baselines—such as DocOwl~1.5—by \increasenoparent{5.5} F1 points, highlighting its robustness to unseen documents. 
These findings show that content‐aware tokenization combined with explicit layout modeling offers a promising direction forward for document understanding. 
Data, source code, and model weights will be made publicly available.
\end{abstract}    
\section{Introduction}
\label{sec:intro}

\begin{figure}[ht]
\centering
\begin{subfigure}[t]{0.48\textwidth}
        \centering
        \includegraphics[width=\linewidth]{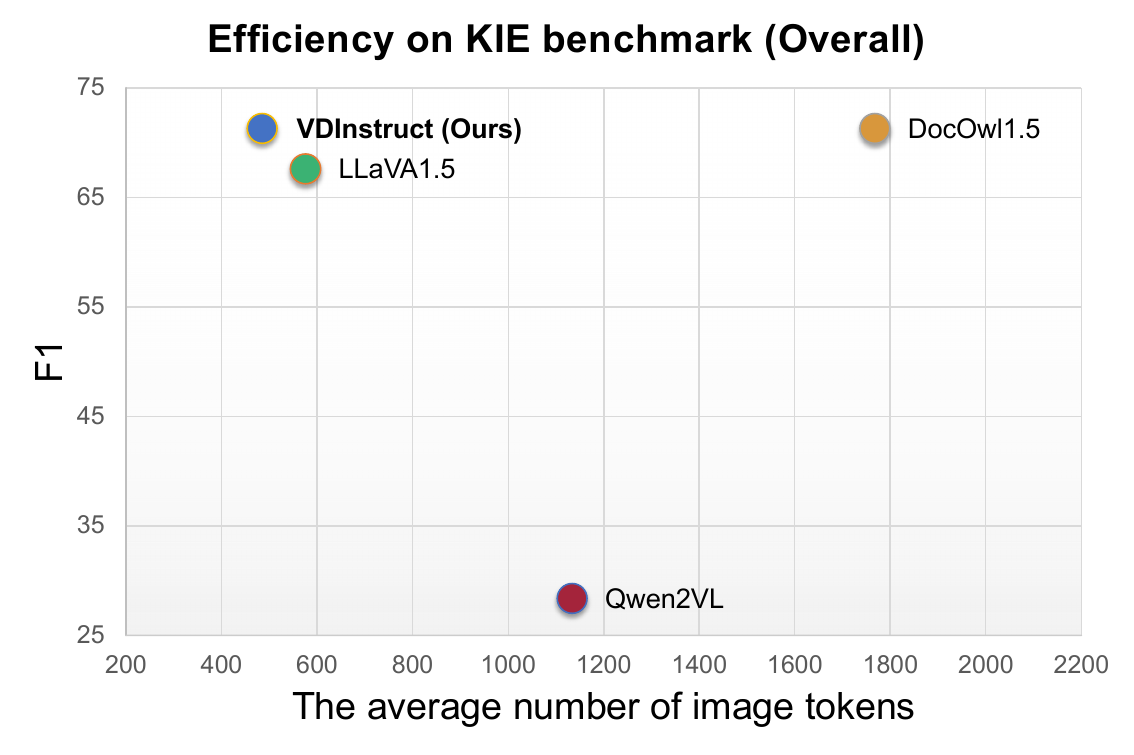}
    \end{subfigure}
    \hfill
    \begin{subfigure}[t]{0.48\textwidth}
        \centering
        \includegraphics[width=\linewidth]{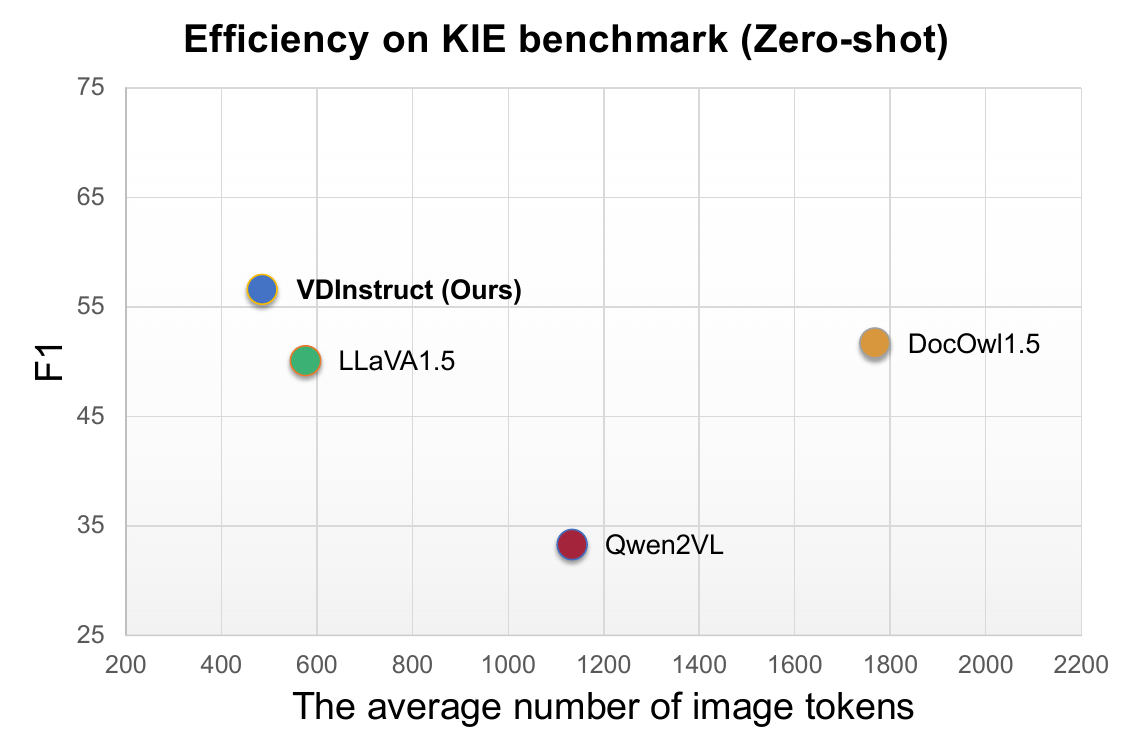}
    \end{subfigure}
\caption{
Image-token consumption (efficiency) and F1 scores on overall and out-of-domain (zero-shot) KIE benchmarks. 
\vdi achieves SOTA F1 scores while using $\approx$500 image tokens per page—$3.6\times$ fewer than DocOwl 1.5~\cite{Hu2024DocOwl15}.
}
\label{fig:image_token_efficiency}
\end{figure}

Visual Document Understanding (VDU) refers to the automated processing of documents (\eg, invoices, receipts, or contracts) that contain both rich visual and textual content.
VDU involves extracting text and pinpointing its exact location on the page, identifying structural components like headings, columns, and charts, and interpreting how these elements interact to convey overall information.
Within this framework, Key Information Extraction (KIE) serves as the core task: it focuses on locating and extracting the most critical fields—such as vendor names, dates, item descriptions, amounts, or contractual clauses—from diverse document types, thereby enabling a wide range of downstream applications~\cite{Wang2024DocLLM}.

However, while existing foundation models for VDU like LayoutLM~\cite{xu2020layoutlm}, Donut~\cite{kim2021donut}, and DocFormer~\cite{appalaraju2021docformer} achieve strong results by fusing text, visual cues, and spatial layout, this performance comes at the cost of extensive domain-specific data annotation and expensive supervised fine-tuning. 
Each new document type requires collecting labeled examples and often fine-tuning a separate model.
In addition, they often struggle to generalize to new document formats without additional annotations, driving the need for more scalable and robust methods that can adapt to diverse layouts and content with minimal supervision.

Multimodal large language models (MLLMs), for example, the LLaVA family~\cite{liu2023visual}, have shown remarkable zero-shot performance on natural-image visual question answering (VQA) via large-scale vision language pretraining combined with instruction tuning. 
Yet, when applied to VDU, MLLMs often fall short, achieving low accuracy on tasks like document information localization and extraction (see~\cref{tab:kie-benchmarks}). 
Unlike natural scenes, documents pack dense, fine-grained text into precise spatial layouts. 
Capturing these requires high-resolution inputs to preserve small tokens and subtle structural cues. 
However, most MLLMs downsample document images (\eg, $336\times336$ for LLaVA 1.5) heavily to fit their context windows, sacrificing critical semantic and spatial information.
To mitigate resolution loss, recent MLLMs, such as DocOwl~1.5~\cite{Hu2024DocOwl15} and Qwen2VL~\cite{wang2024qwen2}, scale image-token counts with pixel resolution by adopting an adaptive cropping module. 
This strategy often splits a single image into multiple patches, encodes each patch separately, and then aggregates all image tokens.
Although this improves accuracy, it is prone to \textbf{token explosion}, which inflates memory usage and risks context-window overflow.
Moreover, naively scaling the number of image tokens with the image size results in \textbf{token redundancy}, since a high-resolution page may contain only several words, and most patches are the background.

To address these limitations, we introduce \vdi, a new multimodal large language model designed for KIE with token-efficient, layout-aware visual encoding. 
Our work introduces two key innovations:
\begin{itemize}
  \item \textbf{Dual Vision Encoder:} We disentangle spatial and semantic processing into two sub-encoders: a \textbf{Spatial Encoder} that is trained with our new layout-aware pretraining objective to model document structure explicitly; a \textbf{Semantic Encoder} extracts fine-grained multimodal (textual + visual) features from high-resolution inputs.
  \item \textbf{Rigorous Benchmark for KIE:}
  We curate the first comprehensive, instruction-based KIE benchmark by unifying eight diverse public document datasets—six for in-domain tuning and two held out for zero-shot robustness testing.
\end{itemize}

Through extensive experiments, we found that: firstly, \vdi is more computationally efficient than state-of-the-art models since it generates only $\approx$500 tokens per page, \textcolor{ForestGreen}{$3.6\times$} fewer than DocOwl 1.5 (see~\cref{fig:image_token_efficiency}). Secondly, despite significantly reducing the number of image tokens, \vdi achieves a SOTA overall performance on KIE benchmarks and sets a new record in zero-shot settings, \increasenoparent{5.5} points gain of F1 scores, compared to DocOwl 1.5~\cite{Hu2024DocOwl15}, the previous state-of-the-art model in VDU (see~\cref{tab:kie-benchmarks}).

\begin{figure*}[ht]
\centering{\includegraphics[width=1.0\textwidth]{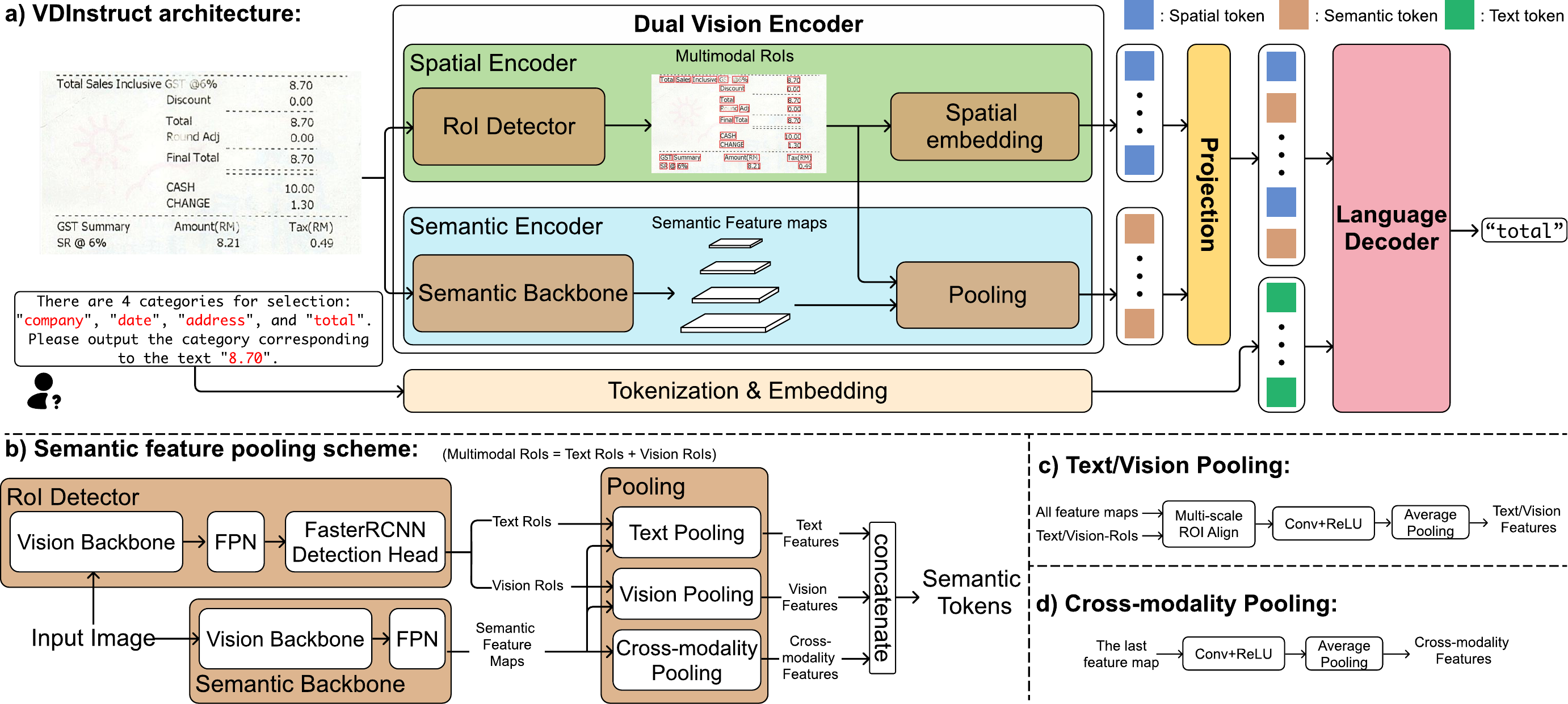}}
\caption{Overview of \vdi. (a) The model architecture. The core module is the dual vision encoder: a spatial encoder to detect text/vision regions (named Multimodal Region of Interest - RoI) and encode them into spatial tokens; a semantic encoder to extract corresponding semantic tokens based on detected RoIs. 
The details of the semantic feature pooling scheme are shown in (b), (c), and (d).}
\label{fig:all_architect}
\end{figure*}

\section{Related Works}
\label{sec:related_work}

\subsection{Multimodal LLMs for Document Understanding}
\label{sec:rw_mllm_doc}

Multimodal Large Language Models (MLLMs) pursue a single, task-agnostic model for document understanding, yet they diverge sharply in balancing vision input and computational cost.  
At one extreme, DocLLM~\cite{Wang2024DocLLM} removes images altogether, supplying the LLM only with OCR tokens and their bounding boxes; its disentangled spatial attention delivers highly efficient reasoning but sacrifices non-textual cues.  
In contrast, mPLUG-DocOwl~\cite{Ye2023DocOwl} couples a vision encoder with an instruction-tuned decoder to achieve fully OCR-free parsing, and version 1.5~\cite{Hu2024DocOwl15} reduces image tokens via horizontal patch merging while preserving SOTA accuracy.
Bridging these philosophies, InstructDoc shows that a single instruction-tuned VL model can zero-shot across 30 public datasets~\cite{tanaka2024instructdoc}, while generalist systems such as Qwen-VL rival much larger models on DocVQA~\cite{mathew2020docvqa} and TextVQA~\cite{singh2019towards} without document-specific tuning~\cite{bai2023qwen}.  
Despite this progress, current MLLMs~\cite{Ye2023DocOwl, ye-etal-2023-ureader, hong2024cogagent, feng2024docpedia} either explode visual tokens at high resolution or blur fine-grained layout by aggressive down-sampling.  
Our dual-encoder architecture resolves this tension: a spatial branch distills layout cues and a semantic branch preserves small text and subtle graphics, together producing 3.6$\times$ fewer tokens than DocOwl 1.5 yet retaining the detail essential for accurate KIE.

\subsection{Key Information Extraction}
\label{sec:rw_kie}

Key Information Extraction (KIE) seeks to automatically surface critical fields from visually rich documents.  
A recent survey by \citet{Rombach2024Deep} highlights both the commercial impact of KIE systems and the rapid research progress driving that impact.  
The modern line of work began with LayoutLM~\cite{xu2020layoutlm}, which first fused token embeddings with their 2-D coordinates; this simple yet powerful idea lifted F1 scores on forms and receipts far beyond those of text-only baselines.  
LayoutLMv2~\cite{Xu2021LayoutLMv2} broadened the signal by injecting image patches and relative 2-D positional biases, yielding robustness to scale and rotation changes, while LayoutLMv3~\cite{Huang2022LayoutLMv3}, StrucTexTv2~\cite{yu2023structextv2}, and SelfDoc~\cite{li2021selfdoc} unified masked pre-training over text and pixels to push accuracy even further across multiple KIE benchmarks.
Complementing this family, DocFormer~\cite{appalaraju2021docformer} used cross-modal attention to explicitly align words with co-located visual features, achieving competitive accuracy with a fraction of the parameters.  
While these works prove KIE benefits from textual, visual, and geometric cues, they still require task-specific fine-tuning and often fail on unseen layouts.  
Our method tackles this gap by coupling instruction tuning with a token-efficient, layout-aware vision encoder, thereby preserving fine-grained spatial detail while enabling true zero-shot extraction.

\section{Methodology}
\label{sec:methodology}

\begin{figure*}[t]
\centering{\includegraphics[width=1.0\textwidth]{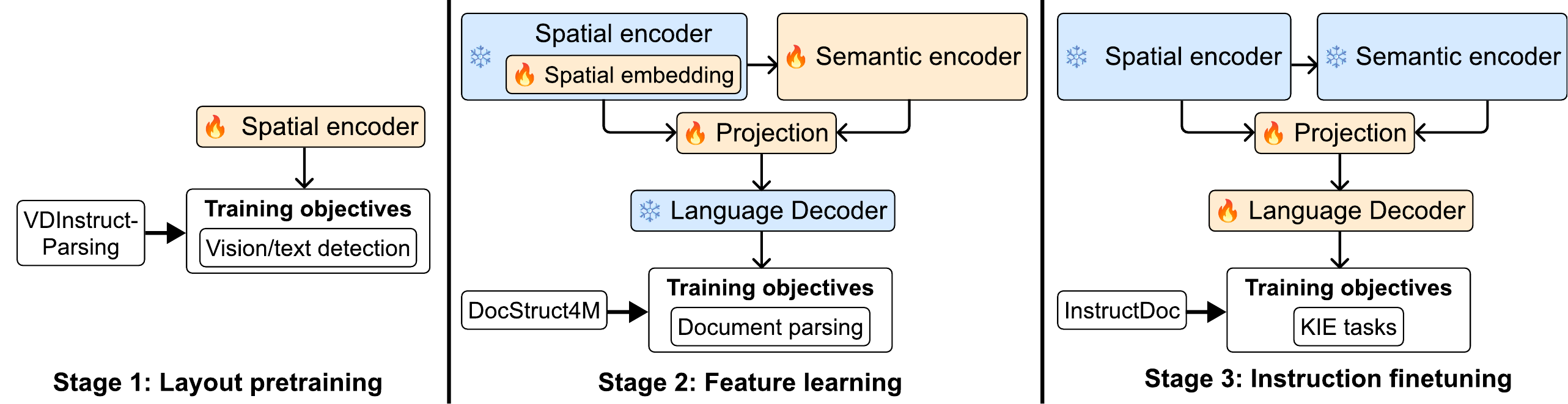}}
\caption{Our proposed 3-stage training paradigm. 
In \textbf{Stage 1}, we train the spatial encoder (without the spatial embedding module) on a detection task using our VDInstruct-Parsing dataset. 
After that, we freeze this module and train the remaining part of the vision encoder on document parsing tasks in \textbf{Stage 2} to learn the meaningful image tokens. 
Finally in \textbf{Stage 3}, we finetune the projection layers and the language decoder on InstructDoc to enable the instruction-following capability.}
\label{fig:training_paradigm}
\end{figure*}

\paragraph{Problem Formulation of Zero-shot KIE}
We focus on the Semantic Entity Recognition (SER) task, a fundamental task in KIE~\cite{jaume2019funsd, wang2023vision, tanaka2024instructdoc}. We frame zero-shot KIE as a conditional sequence generation problem: given a document image \(I\) and a question \(q\) specifying a text segment \(t\) from the document image with a set of all possible entity labels \(E_\text{seen}\) (\eg, invoice date), the model must output the corresponding entity label \(e \in E_\text{seen}\) of the given text segment directly. 
Formally, we learn a mapping: $(I, q(t, E_\text{seen}))\;\longmapsto\;e$. 
During zero-shot testing, models need to process a set of novel entity labels \(E_\text{unseen}\), which are not available during training time. 
This setting requires models not only to localize the given text segment in the input image but also to understand its relation to the unseen entity labels.

\subsection{Architecture of \vdi}

In processing visual documents, semantic content and spatial layout carry distinct information and require specialized modeling.
Most methods rely on a single vision encoder to identify small regions and understand their content, creating a trade-off: either the page is divided into too many patches, leading to token explosion~\cite{Hu2024DocOwl15, wang2024qwen2, li2024monkey}; or regions are merged excessively, leading to the loss of important layout details~\cite{liu2023visual, bai2023qwen}.

To address this, \vdi adopts a dual vision encoder architecture: a spatial encoder that precisely detects and embeds Multimodal Regions of Interest (ROIs), including textual elements (\eg, words, headers) and visual elements (\eg, figures, charts); and a semantic encoder that extracts visual–textual features from each detected ROIs.
By decoupling region localization from feature extraction, each encoder can be trained with objectives tailored to its role, enabling \textbf{content-aware tokenization} (\ie, allocating tokens only to informative regions and filtering out redundant areas) that minimizes redundancy while preserving critical layout cues. 
\cref{fig:all_architect} shows the overall architecture of \vdi. Our model builds on the standard MLLM architecture, which comprises a vision encoder, a projection layer, and a language decoder, but replaces the single vision encoder with a dual vision encoder (see \cref{fig:all_architect}a). 
The \emph{spatial encoder} first detects Multimodal ROIs and converts each ROI into a spatial token that encodes its geometry. In parallel, the \emph{semantic encoder} processes those same ROIs to extract fine-grained content features, producing semantic tokens that capture local meaning. 
We then interleave spatial and semantic tokens by region in reading order, project this structured sequence into the language decoder’s embedding space, and concatenate it with the instruction tokens before decoding. 

\noindent
\textbf{The spatial encoder} is responsible for detecting and encoding layout-specific regions in document images.
As illustrated in ~\cref{fig:all_architect}a, it contains two key components: a RoI detector, which detects text and vision regions in the input image; and a spatial embedding module constructed with a simple linear layer to project the detected ROIs into spatial tokens.
The RoI detector (\cref{fig:all_architect}b) follows the Faster R-CNN architecture~\cite{ren2015faster}: a vision backbone (Swin Transformer v2~\cite{liu2022swin}) feeds into a Feature Pyramid Network (FPN)~\cite{lin2017feature} to address the scale imbalance between small text regions and larger visual objects.
The resulting feature maps are processed by a standard Faster R-CNN detection head to produce \(N\) region proposals (ROIs), each classified as a text-ROI or vision-ROI. 
To preserve global context, we append one additional ROI spanning the entire image,
yielding \(N+1\) total regions. 
Finally, each ROI’s bounding box is projected via a linear layer (the spatial embedding module) into a \(d\)-dimensional vector, producing (\(N+1\)) spatial tokens in \(\mathbb{R}^{(N+1)\times d}\) that precisely encode the geometry of every detected region.

\noindent
\textbf{The semantic encoder }extracts fine-grained semantic tokens based on the \(N\) multimodal ROIs detected by the spatial encoder. 
As shown in~\cref{fig:all_architect}a, it comprises a semantic backbone and a pooling module. 
The semantic backbone consists of a vision backbone (Swin Transformer v2~\cite{liu2022swin}) followed by an FPN~\cite{lin2017feature}, which produces multi-scale feature maps. 
The pooling module takes the multimodal RoIs from the spatial encoder and the feature maps from the semantic backbone. This module contains three parallel submodules: \textit{text pooling} for textual ROIs, \textit{vision pooling} for visual ROIs, and \textit{cross-modality pooling} to capture global feature vectors (see~\cref{fig:all_architect}b).
\cref{fig:all_architect}c shows the text and vision pooling modules, each comprising a multi-scale ROI Align operation, a convolutional layer with ReLU~\cite{nair2010rectified} activation, and adaptive average pooling.
Each text ROI is pooled into a fixed-size feature map of shape \((s_t \times s_t, d)\), while each vision ROI is pooled into \((s_v \times s_v, d)\).
These are then flattened into sequences of length \(s_t^2\) or \(s_v^2\), respectively. 
In addition to region-specific features, we employ a cross-modality pooling module (\cref{fig:all_architect}d) to capture holistic content from the entire image. 
This global representation is obtained by applying convolution, ReLU~\cite{nair2010rectified} activation, and adaptive average pooling to the final backbone feature map, producing a feature map of shape \((s_g \times s_g, d)\), which is then flattened to form \(s_g^2\) cross-modality features. 
These features help mitigate the effect of imperfect detection and retain comprehensive context across modalities. 
Finally, all extracted features are concatenated to form the semantic tokens.

\subsection{Training}
We adopt a three-stage training paradigm (\cref{fig:training_paradigm}) that incrementally builds \textsc{{VDInstruct}}’s abilities, from layout detection $\rightarrow$ semantic understanding $\rightarrow$ instruction following, while allowing each component to specialize and minimizing interference between objectives.

\noindent
\textbf{Layout pretraining (Stage 1).} The first stage of training aims to teach the spatial encoder to detect multimodal ROIs in document images. Unlike traditional OCR systems, which focus on recognizing and transcribing text, this stage is concerned solely with \textit{region-level detection} (see~\cref{app:roi-detector}). The training objective at this stage is identical to standard object detection.
To support robust pretraining, we construct a dataset named \textbf{VDInstruct-Parsing}, which aggregates document images from seven public sources: AI2D~\cite{kembhavi2016diagram}, DocLayNet~\cite{pfitzmann2022doclaynet}, DocBank~\cite{li-etal-2020-docbank}, SciCap~\cite{hsu-etal-2021-scicap-generating}, ScienceQA~\cite{lu2022learn}, KLC~\cite{borchmann2021due}, and PWC~\cite{borchmann2021due}.
These datasets span various document types, including scientific papers, academic diagrams, forms, and educational content. 
For \textit{text annotations}, we leverage existing OCR outputs when available; otherwise, we employ the Azure OCR engine~\cite{azureocr}. 
For \textit{vision annotations}, we annotate rich-visual areas, such as natural objects, figures, and charts, as visual ROIs (see~\cref{app:vdinstruct-parsing}).

\noindent
\textbf{Feature learning (Stage 2).} This stage aims to extract meaningful visual representations from the detected ROIs by training the semantic encoder and the spatial embedding layer of the spatial encoder. 
We adapt the document parsing task from DocOwl~1.5 and DocStruct4M dataset~\cite{Hu2024DocOwl15}, formatted in a VQA style with simple parsing instructions. During training, the language decoder is kept frozen, and only the semantic encoder, spatial embedding layer, and projection layer are updated. 
This allows the model to align visual features with text-space embeddings without affecting the linguistic capacity of the decoder. 
The training objective combines document structure parsing and fine-grained text localization, encouraging the model to retain content semantics and layout information in its vision tokens.

\noindent
\textbf{Instruction tuning (Stage 3).} The third stage focuses on instruction fine-tuning, aiming to equip the language decoder with the ability to understand and execute more complex tasks, particularly within the VDU domain. 
For this purpose, we utilize the InstructDoc dataset~\cite{tanaka2024instructdoc}, which is specifically designed to evaluate the zero-shot capabilities of VLMs on document intelligence tasks. 
The training data of InstructDoc incorporates approximately 120,000 images, including samples from six KIE datasets (Deepform~\cite{borchmann2021due}, DocILE~\cite{vsimsa2023docile}, PWC~\cite{borchmann2021due}, KLC~\cite{borchmann2021due}, SROIE~\cite{huang2019icdar2019}, and Wildreceipt~\cite{sun2021spatial}), while FUNSD~\cite{jaume2019funsd} and CORD~\cite{park2019cord} are held out for zero-shot evaluation. 
A key advantage of InstructDoc is its provision of a unified format for converting these diverse KIE datasets into a consistent instruction-following structure, directly matching the scope of our work. 
To specifically isolate the contribution of our vision encoder design to zero-shot KIE performance, we deliberately avoid incorporating large-scale, general-purpose instruction-following datasets often used in VLM training.
\section{Experiments}
\label{sec:experiments}

\subsection{Training Setup}

For the vision backbone, we select Swin Transformer (SwinB-v2)~\cite{liu2022swin} due to its effective balance between performance and computational cost. 
We also experiment with ResNet-50~\cite{he2016deep}. 
To control token granularity, we use different pooling sizes for each modality to produce a similar number of image tokens as LLaVA 1.5~\cite{liu2024improved}, which uses the same language decoder as \vdi. 
While larger pooling sizes can help preserve more information, matching LLaVA 1.5’s token count allows us to attribute performance gains to the quality of our vision tokens rather than sheer quantity.
Following this heuristic, \textit{text ROIs} are pooled to a $(1,1)$ size, yielding one token per region; \textit{vision ROIs} to $(4,4)$, producing 16 tokens; and \textit{cross-modality features} to $(8,8)$, contributing 64 high-level contextual tokens. 
This setting results in approximately 500 image tokens per input, comparable to LLaVA 1.5’s 576. 
Please see~\cref{app:image-token-count} for the detailed derivation of token counts.

We plug this vision backbone into the LLaVA framework, utilizing Vicuna v1.5 7B~\cite{vicuna2023} as the large language model. 
The projection module connecting vision and language features follows the LLaVA v1.5~\cite{liu2024improved} design (2 linear layers $\rightarrow$ a GeLU activation~\cite{hendrycks2016gaussian}) but is trained from scratch in our pipeline. 
Input images are always resized to $1024 \times 1024$, normalized with ImageNet statistics~\cite{he2016deep} without augmentations.
Implementation details and training configurations are provided in~\cref{app:implement-details}.

\begin{table*}[t]
\caption{
Evaluation (F1 scores and image tokens) on KIE datasets. 
"V", "T", and "L" denote input modalities: image, text, and bounding boxes (text and bounding boxes are extracted by an external OCR tool), respectively. Models with “*” are imported from InstructDoc~\cite{tanaka2024instructdoc}. \vdi shows the SOTA results while significantly reducing the image tokens.
}
\label{tab:kie-benchmarks}
\centering 
\resizebox{\textwidth}{!}{%
\begin{tabular}{|l|cc|ccccccc|ccc|c|}
\hline
 &  &  & \multicolumn{7}{c|}{\textbf{In-domain}} & \multicolumn{3}{c|}{\textbf{Out-of-domain (zero-shot)}} & \textbf{Overall} \\
\textbf{Model} & \textbf{\#Param} & \textbf{Modal} & Deepform & DocILE & KLC & PWC & SROIE & Wildreceipt & \textbf{Avg} & FUNSD & CORD & \textbf{Avg} & \textbf{Avg} \\
\hline
LLaVAR~\cite{zhang2023llavar} & 13.3B  & V & 0.7 & 0.0 & 0.1 & 0.6 & 0.2 & 0.0 & 0.3 & 0.6 & 0.0 & 0.3 & 0.3 \\
MiniGPT-v2~\cite{chen2023minigpt} & 7B  & V & 6.3 & 5.2 & 0.3 & 15.3 & 49.8 & 0.3 & 12.9 & 4.4 & 18.9 & 11.7 & 12.6 \\
InstructBLIP~\cite{dai2023instructblip} & 3.4B  & V & 0.0 & 0.0 & 0.0 & 0.9 & 0.0 & 0.0 & 0.2 & 0.0 & 0.0 & 0.0 & 0.1 \\
BLIP-2~\cite{li2023blip} & 3.4B & V & 0.0 & 0.0 & 0.0 & 0.0 & 0.0 & 0.0 & 0.0 & 0.0 & 0.0 & 0.0 & 0.0 \\
Monkey~\cite{li2024monkey} & 9.8B & V & 19.0 & 1.3 & 19.4 & 11.2 & 70.5 & 1.8 & 20.5 & 13.1 & 10.4 & 11.8 & 18.3 \\
TextMonkey~\cite{liu2024textmonkey} & 9.7B & V & 19.3 & 0.2 & 3.2 & 14.2 & 34.5 & 1.1 & 12.1 & 8.0 & 0.0 & 4.0 & 10.1 \\
DocLayLLM~\cite{liao2024doclayllm} & 8B & VTL & 60.4 & 13.8 & 72.9 & 17.1 & 89.0 & 48.8 & 50.3 & 19.9 & 48.9 & 34.4 & 46.4 \\
LLaVA 1.5~\cite{liu2024improved} & 7.3B & V & 34.8 & 13.7 & 41.8 & 14.3 & 71.0 & 12.9 & 31.4 & 10.9 & 31.2 & 21.1 & 28.8 \\
DocOwl 1.5~\cite{Hu2024DocOwl15} & 8.1B & V & 36.2 & 5.5 & 23.3 & 19.7 & 57.5 & 5.0 & 24.5 & 20.6 & 12.6 & 16.6 & 22.6 \\
\hline
\rowcolor[gray]{0.9} \multicolumn{14}{|l|}{\textbf{Finetuned on InstructDoc:}} \\
BLIP-2~\cite{li2023blip}* & 3.4B & VT & - & - & - & - & - & - & - & 26.0 & 33.8 & 29.9 & - \\
InstructDr~\cite{tanaka2024instructdoc}* & 3.4B & VTL & - & - & - & - & - & - & - & 38.2 & 46.0 & 42.1 & - \\
Qwen2VL~\cite{wang2024qwen2} & 7B & V & 49.1 & 21.7 & 22.1 & 4.1 & 25.0 & 22.5 & 24.1 & 30.3 & 36.2 & 33.3 & 26.4 \\
LLaVA 1.5~\cite{liu2024improved} & 7.3B & V & 80.1 & 71.3 & 86.3 & \textbf{20.2} & 99.8 & 82.8 & 73.4 & 40.1 & 60.1 & 50.1 & 67.6 \\
DocOwl 1.5~\cite{Hu2024DocOwl15} & 8.1B & V & \textbf{96.3} & \textbf{78.7} & \textbf{86.6} & 20.0 & 99.8 & \textbf{85.6} & \textbf{77.8} & 45.5 & 57.8 & 51.7 & 71.3 \\
\textbf{\vdi} & 7.2B & V & 93.3 & 74.2 & \textbf{86.6} & 20.0 & \textbf{99.9} & 83.1 & 76.2 & \textbf{50.7} & \textbf{63.6} & \textbf{57.2} & \textbf{71.4} \\
\hline
\rowcolor[gray]{0.9} \multicolumn{14}{|l|}{\textbf{Number of image tokens:}} \\
DocOwl 1.5~\cite{Hu2024DocOwl15} & 8.1B & V & 1799 & 1802 & 1799 & 1799 & 1767 & 1508 & 1746 & 1799 & 1868 & 1834 & 1768 \\
\textbf{\vdi} & 7.2B & V & 808 & 488 & 219 & 1027 & 377 & 282 & 534 & 488 & 192 & 340 & 485 \\
 & & & (\textbf{2.2$\times$$\downarrow$}) & (\textbf{3.7$\times$$\downarrow$}) & (\textbf{8.2$\times$$\downarrow$}) & (\textbf{1.8$\times$$\downarrow$}) & (\textbf{4.7$\times$$\downarrow$}) & (\textbf{5.3$\times$$\downarrow$}) & (\textbf{3.3$\times$$\downarrow$}) & (\textbf{3.7$\times$$\downarrow$}) & (\textbf{9.7$\times$$\downarrow$}) & (\textbf{5.4$\times$$\downarrow$}) & (\textbf{3.6$\times$$\downarrow$}) \\
\hline
\end{tabular}}
\end{table*}

\subsection{Evaluation Metrics}

We measure performance at each training stage using established metrics.  
For ROI detection, we report mean Average Precision (mAP), a standard object‐detection metric~\cite{lin2014microsoft}.
For KIE benchmarks, we use the F1 score~\cite{jaume2019funsd, wang2023vision}, which balances precision and recall, as our primary metric.  
We evaluate under two conditions: \textbf{In‐domain}: test splits of the six KIE datasets used in InstructDoc training (Deepform, DocILE, KLC, PWC, SROIE, Wildreceipt). \textbf{Out‐of‐domain}: held‐out FUNSD and CORD datasets to assess zero‐shot generalization.
This setup lets us quantify both supervised accuracy and true zero‐shot performance.  

\subsection{Findings}
\noindent
\textbf{Spatial encoder achieves high multimodal ROI detection accuracy.} 
Since the spatial encoder is responsible for capturing document structure via multimodal ROIs, its detection accuracy is the driving force for preserving layout information and underpins \textsc{{VDInstruct}}’s overall performance. We quantify this capability using mean Average Precision (mAP), the standard object‐detection metric. 
With SwinB-v2~\cite{liu2022swin} as the backbone, our ROI detector achieves an mAP of 0.611 and an AP@50 of 0.818, showing strong localization accuracy. 
Class‐specific APs—0.605 for text regions and 0.618 for vision regions—confirm balanced performance across modalities. 
Please see ~\cref{tab:detection} for more ablation studies on different vision backbone choices, and~\cref{app:detect-qualitative} for the qualitative results of RoI detection.

\begin{figure*}[t]
\centering{\includegraphics[width=0.98\textwidth]{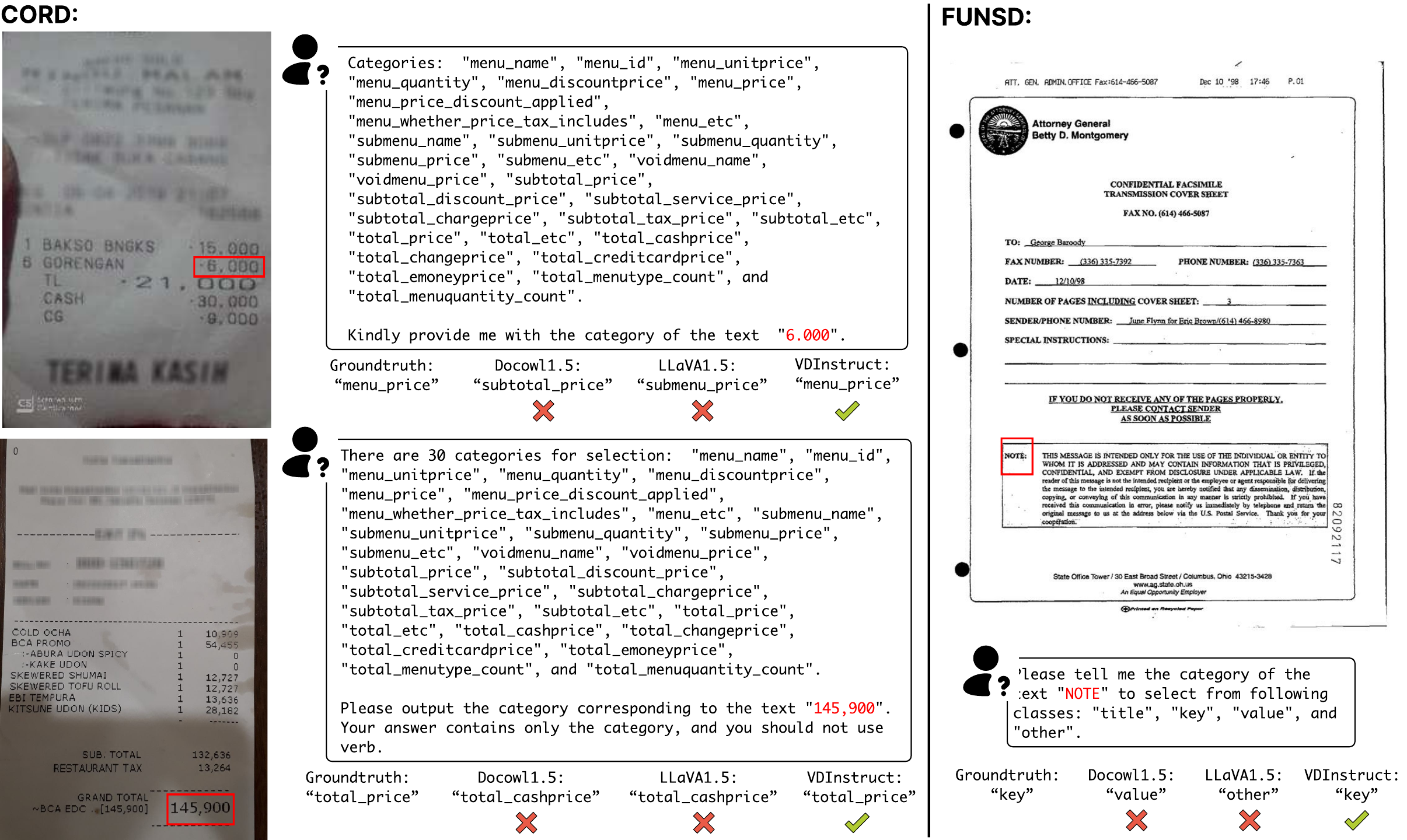}}
\caption{Qualitative results on the out-of-domain datasets. The \textcolor{red}{red} bounding boxes indicate the text segments mentioned in the questions. \vdi can correctly predict the labels of the selected text segments, whereas DocOwl 1.5 and LLaVA 1.5 cannot.}
\label{fig:qualitative_results}
\end{figure*}

\noindent
\textbf{VDInstruct achieves high accuracy on KIE benchmarks.}
\cref{tab:kie-benchmarks} reports KIE results for \vdi and our selected baselines. 
We restrict comparisons to models with publicly available pretrained checkpoints before instruction fine-tuning, ensuring that performance differences reflect the vision encoder design rather than extra instruction-tuning data.
We fine-tune 3 models, including DocOwl 1.5~\cite{Hu2024DocOwl15}, LLaVA 1.5~\cite{liu2024improved}, and Qwen2VL~\cite{wang2024qwen2}, on the same InstructDoc dataset used to train \vdi.
BLIP-2~\cite{li2023blip} and InstructDr~\cite{tanaka2024instructdoc} are imported from InstructDoc Tanaka \etal ~\cite{tanaka2024instructdoc}.
Additionally, we include reference models trained with their own instruction-following data to showcase the performance without fine-tuning on InstructDoc, such as MiniGPT-v2~\cite{chen2023minigpt} and Monkey~\cite{li2024monkey}.
Testing on the in-domain setting (\cref{tab:kie-benchmarks}), \vdi achieves an average F1 of 76.2, ranking second-best on four out of six datasets and achieving the best score on KLC and SROIE. 
Although \vdi slightly reduces accuracy on supervised datasets, it achieves notable improvements in out-of-domain performance, suggesting enhanced generalization for KIE tasks.
To be specific, in the zero-shot setting, our model achieves SOTA results with an average F1 of 57.2, outperforming all other baselines by a large margin. 
Compared to DocOwl 1.5, \vdi outperforms this model by a significant gap, \increasenoparent{5.5} points. 
Our model also consistently beats Qwen2VL and LLaVA 1.5 on all datasets.
Compared to LLaVA 1.5, which shares the same language decoder but uses a different vision pipeline, \vdi shows a performance gain of \increasenoparent{7.1} points, indicating the efficiency of our dual vision encoder design.
Furthermore, despite not using text recognition systems, \vdi surpasses both OCR-free (LLaVA 1.5, DocOwl 1.5, Qwen2VL) and OCR-based models (InstructDr). 
Finally, our model achieves the best overall average scores, 71.4 points.
In \cref{fig:qualitative_results}, we demonstrate the qualitative results of \vdi and its counterparts on zero-shot samples. 
These findings highlight the effectiveness of our dual vision encoder and layout-aware training in achieving robust KIE performance.

\noindent
\textbf{Efficient image tokenization: 3.6$\times$ fewer tokens without accuracy trade-off.} Beyond accuracy, we also measure how efficiently each model encodes document images by reporting the \textbf{average number of image tokens} per document. 
Efficiency is critical: every extra token incurs risks overflowing the decoder’s limited context window, directly impacting computational cost and scalability~\cite{yoon-etal-2024-eyes}.
In~\cref{fig:image_token_efficiency} and~\cref{tab:kie-benchmarks}, \vdi produces approximately 500 image tokens per document on average, which is about \textcolor{ForestGreen}{3.6} times fewer than DocOwl 1.5. 
Despite this significant reduction, \vdi achieves SOTA performance across KIE datasets, demonstrating that our dual-encoder design effectively preserves essential semantic and layout information without generating redundant tokens. 
Compared to LLaVA 1.5, which shares the same architecture for the language decoder and projection layer with \vdi, our model consistently achieves better KIE performance despite producing a similar number of image tokens.
This result suggests that while other counterparts scale image tokens with input resolution, \vdi scales tokens with the document's content, enabling more effective and efficient image encoding.

\begin{table}
\caption{F1 scores of \vdi when using different semantic token modality combinations. 
\textsc{Cross}, \textsc{Text}, \textsc{Vision} correspond to cross-modality, textual-modality, and visual-modality semantic tokens, respectively. 
\vdi achieves the best performance on KIE benchmarks when using all semantic tokens (Cross+Text+Vision). 
We add LLaVA 1.5 for a better comparison.
}
\label{tab:ablation-token-modality}
\resizebox{\linewidth}{!}{
\begin{tabular}{|l|c|c|c|}
\hline
\textbf{Modality} & \textbf{Avg (ID)} & \textbf{Avg (OOD)} & \textbf{Overall}\\
\hline
Cross & 71.2 & 55.3 & 67.2 \\
Cross+Vision & 71.2 & 57.1 & 67.6 \\
Cross+Text & 76.1 & 56.3 & 71.2 \\
\textbf{Cross+Text+Vision} & \textbf{76.2} & \textbf{57.2} & \textbf{71.4} \\
\hline
\hline
LLaVA 1.5~\cite{liu2024improved} & 73.4 & 50.1 & 67.6 \\
\hline
\end{tabular}
}
\end{table}

\noindent
\textbf{Complementary token modalities deliver best performance.} We also investigate the impact of different modalities of the \emph{semantic} tokens produced by the dual vision encoder. 
In~\cref{tab:ablation-token-modality}, we start with using only cross-modality tokens (\textsc{Cross}), then gradually add textual-modality (\textsc{Text}) and visual-modality tokens (\textsc{Visual}). 
In this experiment, we do not re-train the entire model but simply ablate out the corresponding tokens in each setting during the inference.
We use LLaVA 1.5 for reference since it shares the same language decoder (Vicuna v1.5 7B) with our model. 
With only cross-modality tokens, \vdi underperforms its counterpart, LLaVA 1.5, although it performs better on out-of-domain benchmarks. 
This is reasonable since image tokens consumed by \vdi in this case are much smaller than LLaVA 1.5 (64 compared to 576). 
When adding visual-modality tokens to the vision encoder's outputs, \vdi achieves the same F1 score with LLaVA 1.5 by increasing its zero-shot accuracy by 1.8 points. 
Pairing cross-modality and text-modality tokens boosts \vdi's performance by a significant gap, from 67.2 to 71.2-underscoring the critical role of fine-grained text segments in document images. 
Finally, the combination of all token modalities achieves the highest scores across all benchmarks, as reported in~\cref{tab:kie-benchmarks}. 
These results demonstrate that the different types of semantic tokens are complementary; each modality contributes unique information that, when combined, yields the strongest overall performance. 

\noindent
\textbf{Spatial tokens contribute meaningfully to KIE performance.}
To assess the contribution of spatial tokens, we conduct an ablation study by removing them from the input to the language decoder while keeping all other components intact. As shown in~\cref{tab:spatial-token}, excluding spatial tokens leads to a performance drop in both in-domain and out-of-domain settings. Specifically, the overall F1 score decreases from 71.4 to 70.7, with a more pronounced decline in zero-shot (out-of-domain) performance. This highlights the role of spatial tokens in encoding layout-specific cues that complement semantic features and support robust generalization across document types.

\begin{table*}[t]
\centering
\begin{minipage}{0.48\linewidth}
    \caption{Comparison of detection performance between ResNet-50 and Swin-B v2 backbones. Swin-B v2 achieves better results across every evaluation metric.}
    \label{tab:detection}
    \centering
    \small
    \begin{tabular}{|l|c|c|c|c|}
    \hline
    \textbf{Backbone} & \textbf{AP} & \textbf{AP50} & \textbf{AP\_text} & \textbf{AP\_vision} \\
    \hline
    ResNet-50~\cite{he2016deep} & 0.443 & 0.614 & 0.561 & 0.325 \\
    SwinB-v2~\cite{liu2022swin} & \textbf{0.611} & \textbf{0.818} & \textbf{0.605} & \textbf{0.618} \\
    \hline
    \end{tabular}
\end{minipage}
\hfill
\begin{minipage}{0.48\linewidth}
    \caption{F1 scores of \vdi with and without spatial tokens. The performance of \vdi drops on both in-domain and out-of-domain when excluding spatial tokens. }
    \label{tab:spatial-token}
    \centering
    \small
    \begin{tabular}{|l|c|c|c|}
    \hline
    \textbf{Configuration} & \textbf{Avg (ID)} & \textbf{Avg (OOD)} & \textbf{Overall Avg} \\
    \hline
    w/ spatial tokens & 76.2 & 57.2 & 71.4 \\
    w/o spatial tokens & 75.8 & 55.2 & 70.7 \\
    \hline
    \end{tabular}
\end{minipage}
\end{table*}

\begin{table*}[t]
\caption{KIE benchmark comparison between different vision backbones. \vdi performs best with SwinB-v2 as backbones for both spatial and semantic encoders.}
\label{tab:kie}
\centering
\resizebox{\textwidth}{!}{%
\begin{tabular}{|l|ccccccc|ccc|c|}
\hline
\textbf{Backbone} & Deepform & DocILE & KLC & PWC & SROIE & WR & \textbf{Avg (ID)} & FUNSD & CORD & \textbf{Avg (OOD)} & \textbf{Overall Avg} \\
\hline
ResNet-50~\cite{he2016deep}  & 79.7 & 63.8 & 84.4 & 20.0 & 99.9 & 78.0 & 71.0 & 38.0 & 63.8 & 50.9 & 66.0 \\
SwinB-v2~\cite{liu2022swin} & 93.3 & 74.2 & 86.4 & 20.0 & 99.9 & 83.1 & \textbf{76.2} & 50.7 & 63.6 & \textbf{57.2}& \textbf{71.4}\\
\hline
\end{tabular}}
\end{table*}

\noindent
\textbf{Stronger vision backbone improves detection and zero-shot KIE.}
To quantify the effect of vision backbone selection on ROI detection and KIE performance, we conducted an ablation comparing ResNet-50~\cite{he2016deep} and SwinB-v2~\cite{liu2022swin} as the shared backbone for both spatial and semantic encoders.
As~\cref{tab:detection} shows, SwinB-v2 dramatically improves ROI localization, raising mAP from 0.443 to 0.611 and AP@50 from 0.614 to 0.818, while also delivering balanced text and vision APs. 
These gains carry over to downstream key information extraction: after fine-tuning on InstructDoc, as shown in~\cref{tab:kie}, \vdi with SwinB-v2 achieves an in-domain F1 of 76.2 and an out-of-domain F1 of 57.2—\increasenoparent{5.2} and \increasenoparent{5.7} points higher than the ResNet-50 variant—resulting in an overall F1 of 71.4 versus 66.0.

\section{Conclusion}
In this paper, we introduce \vdi, a multimodal LLM for Key Information Extraction that decouples spatial layout detection and semantic feature extraction via a dual-vision encoder.
By leveraging \emph{content-aware tokenization}—allocating tokens to the most informative regions—and a three-stage training paradigm (layout pretraining, feature learning, and instruction tuning), \vdi shows greater vision encoding efficiency while achieving strong generalization in zero-shot settings and maintaining SOTA performance on KIE tasks.

Despite these strong results on KIE benchmarks, \textsc{{VDInstruct}}’s performance depends heavily on the accuracy of the spatial encoder’s predictions. Incorrect or overlapping bounding box detections may lead to redundant tokens and affect overall performance. Future work could remedy this problem by expanding the training dataset to include more diverse document types and resolutions to improve the model's robustness further.
{
    \small
    \bibliographystyle{ieeenat_fullname}
    \bibliography{main}
}

\clearpage
\setcounter{page}{1}
\maketitlesupplementary
\appendix

\section{\vdi vs. existing region‐based VDU models}
\label{app:roi-detector}

While decoupling region detection from feature extraction has precedent in document understanding literature~\cite{xu2020layoutlm, li2021selfdoc, gu2021unidoc, powalski2021going}, \vdi's dual‐vision encoder differs from prior work in two fundamental ways:

\begin{itemize}
\item \textbf{OCR-free end‐to‐end pretrained region detector covering both text and vision regions.}  
  Prior works such as LayoutLM~\cite{xu2020layoutlm}, SelfDoc~\cite{li2021selfdoc}, and UDoc~\cite{gu2021unidoc} employ an external OCR tool to get the Region of Interest (ROI).
  While this approach can preserve the textual information on the document, it neglects the importance of visual cues.
  \vdi instead builds its spatial encoder from a ResNet-50 or Swin Transformer v2 backbone with an FPN feeding into a Faster R-CNN detection head, which is pretrained end-to-end on a heterogeneous suite of document parsing datasets (Stage 1: Layout pretraining) to directly optimize text- and vision-ROI detection for document layouts, eliminating the need for an external OCR tool. 
  Moreover, our ROI detector in the spatial encoder differs from OCR tools since it only focuses on region localization, not text recognition.
  Each detected bounding box is then projected via a learned linear layer into a \(d\)-dimensional spatial token, enabling region proposals to adapt to document‐centric structures rather than generic object categories.

\item \textbf{Task‐aligned semantic encoder rather than pooled region features.}  
  Prior methods extract a single fixed vector per region and concatenate it with text embeddings ~\cite{xu2020layoutlm, li2021selfdoc, gu2021unidoc, powalski2021going}. 
  \vdi’s semantic encoder instead applies a high-resolution backbone (ResNet-50 or Swin Transformer v2 with FPN) to produce multi-scale feature maps, then uses three specialized pooling modules—text pooling for text ROIs, vision pooling for visual ROIs, and cross-modality pooling on the global feature map, each involving multi-scale ROI Align, Conv+ReLU, and adaptive average pooling.
  Additionally, each detected ROI is pooled into a different size (based on its modality) and flattened to a set of feature vectors rather than one fixed vector per region.
  The resulting semantic tokens are concatenated to form rich, context-enriched region representations that preserve fine-grained cues (\eg, fonts, textures) and holistic layout information often lost by static RoI-pooled vectors.  

\end{itemize}

These design choices—OCR-free end‐to‐end trainable region proposals and multi‐scale semantic alignment—distinguish \vdi’s dual‐vision encoder from existing region‐based document models and underpin its superior performance on both in‐domain and zero‐shot tasks.

\section{VDInstruct-Parsing statistics}
\label{app:vdinstruct-parsing}
From each dataset, we randomly sample up to 50{,}000 images from the training split to construct a balanced and diverse training set. To maintain data quality and reduce noise, we exclude images that contain more than 1{,}000 bounding boxes. A similar sampling strategy is applied to create the test set using the original test or validation splits (with a maximum 1{,}000 samples per dataset). For \textit{vision annotations}, we follow dataset-specific rules: AI2D uses the \texttt{blob} attribute, DocBank uses the \texttt{figure} class, and DocLayNet uses the \texttt{picture} class. For datasets such as SciCap and ScienceQA, where graphical content is often embedded as standalone images, we place these figures on a white background to synthesize a consistent visual region. In cases like KLC and PWC, where predefined visual annotations are lacking, we apply a pretrained document layout analysis model~\cite{Huang2022LayoutLMv3} to infer vision regions, taking advantage of the visual similarity between these datasets and the model’s original training data. ~\cref{tab:parsing-dataset-stats} and~\cref{tab:parsing-dataset-boxes} summarize the sample and bounding box distributions across the seven datasets.

\begin{table*}[ht]
\caption{Sample distribution in the VDInstruct-Parsing dataset across training and test sets.}
\label{tab:parsing-dataset-stats}
\centering
\begin{tabular}{|l|cccccccc|}
\hline
 & AI2D & DocLayNet & DocBank & SciCap & ScienceQA & KLC & PWC & \textbf{Total} \\
\hline
\textbf{Train} & 3{,}921 & 48{,}179 & 49{,}301 & 49{,}997 & 6{,}218 & 1{,}729 & 208 & 159{,}553\\
\textbf{Test} & 982 & 975 & 990 & 1{,}000 & 1{,}000 & 440 & 63 & 5{,}450 \\
\hline
\end{tabular}
\end{table*}

\begin{table*}[ht]
\caption{Bounding box statistics in the \vdi-Parsing dataset.}
\label{tab:parsing-dataset-boxes}
\centering
\begin{tabular}{|l|c|c|c|}
\hline
 & \textbf{Text Boxes} & \textbf{Vision Boxes} & \textbf{Total}  \\
\hline
\textbf{Train} & 46{,}403{.}1K & 113{.}1K & 46{,}516{.}2K \\
\textbf{Test} & 983{.}0K & 7{.}6K & 990{.}6K \\
\hline
\end{tabular}
\end{table*}

\section{Training details of \vdi}
\label{app:training_details}

\subsection{The number of image tokens}
\label{app:image-token-count}

\begin{figure*}[t]
\centering{\includegraphics[width=1.0\textwidth]{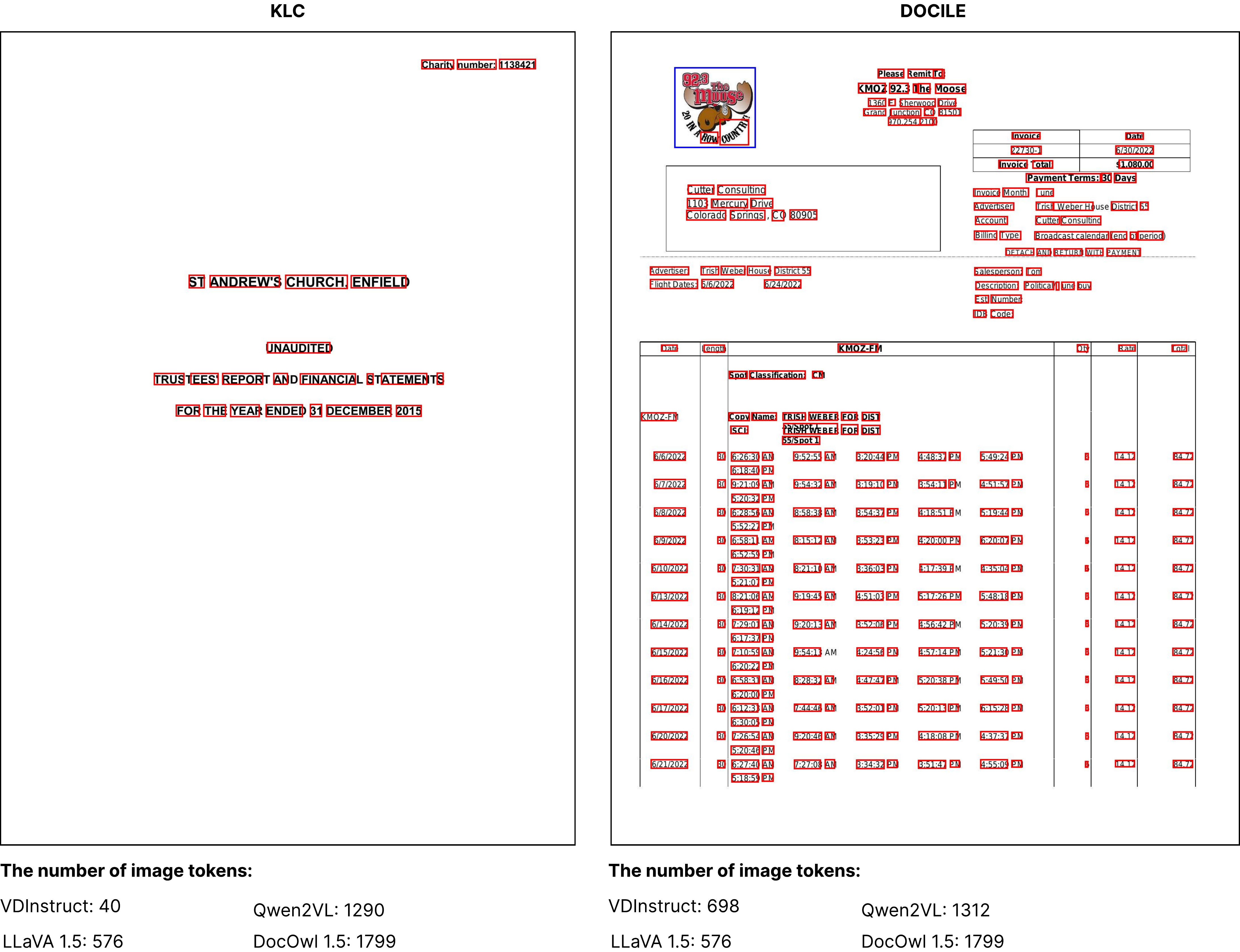}}
\caption{The number of image tokens encoded by \vdi and its counterparts. The red and blue boxes cover the detected text and vision ROIs, respectively. Our dual vision encoder scales the number of image tokens with the detected contents, minimizing token redundancy and information loss.}
\label{fig:tokens_quanlitative}
\end{figure*}

The total number of image tokens produced by \vdi scales with the content of the document rather than its pixel resolution, which helps minimize both token redundancy and information loss. 
Let \(N_t\) and \(N_v\) denote the number of text ROIs and vision ROIs, respectively. 
Then, the total number of detected regions is \(N = N_t + N_v\). 
The number of spatial tokens is:
\begin{equation}
N_{\text{spatial\_tokens}} = N_t + N_v + 1
\label{eq:n_spatial_tokens}
\end{equation}
where the one extra token corresponds to the bounding box covering the entire image. 

Using pooling, each text ROI contributes \(s_t^2\) features (tokens) and each vision ROI contributes \(s_v^2\) features, while the global branch adds \(s_g^2\) cross-modality features. 
Thus, the total number of semantic tokens is given by:
\begin{equation}
N_{\text{semantic\_tokens}} = N_t \cdot s_t^2 + N_v \cdot s_v^2 + s_g^2
\label{eq:n_semantic_tokens}
\end{equation}

Finally, the total number of image tokens, denoted as \(N_{\text{image\_tokens}}\), is:
\begin{multline}
N_{\text{image\_tokens}} = N_{\text{spatial\_tokens}} + N_{\text{semantic\_tokens}} \\
= (N_t + N_v + 1) + \left(N_t \cdot s_t^2 + N_v \cdot s_v^2 + s_g^2\right)
\label{eq:n_image_tokens}
\end{multline}

This content-aware token allocation allows \vdi to capture all necessary details while keeping the token count low, regardless of the input’s overall resolution.

To determine the appropriate pooling sizes for the semantic encoder, we analytically estimate the number of image tokens generated per input image. 
We set \( s_t = 1 \), reflecting the word-level granularity of text RoIs, such that each text RoI is encoded into a single token. To determine \( s_v \) and \( s_g \), we analyze the average size of vision RoIs relative to the input image resolution (1024×1024), finding that the average size of vision RoIs is roughly half the image size. Based on this, we set \( s_g \approx 2 \cdot s_v \).

Given the average number of RoIs per image (\( N_t \approx 293.37 \), \( N_v \approx 0.71 \)) (see~\cref{tab:parsing-dataset-stats} and~\cref{tab:parsing-dataset-boxes}) and the goal of maintaining token efficiency comparable to LLaVA 1.5~\cite{liu2024improved} (our direct counterpart, which uses 576 vision tokens), we aim to constrain the total number of image tokens to the range \([576 - 100, 576 + 100]\). By solving:

\begin{equation}
476 < (N_t + N_v + 1) + (N_t + (N_v + 4) \cdot s_v^2) \leq 676
\end{equation}
we find that \( s_v \leq 4 \) meets this constraint. To minimize information loss, we select the maximum feasible pooling resolution: \( s_v = 4 \), and consequently set \( s_g = 8 \). This results in 1 token per text RoI, 16 tokens per vision RoI, and 64 global tokens per image, balancing efficiency and expressiveness. Please see~\cref{fig:tokens_quanlitative} for examples on KLC and DocILE datasets.

\subsection{Implementation details}
\label{app:implement-details}

All our implementations use PyTorch~\cite{paszke2017automatic} with Python 3.10. All training uses AdamW~\cite{loshchilov2017decoupled} with cosine learning rate scheduler~\cite{loshchilov2016sgdr}. We also leverage DeepSpeed~\cite{rasley2020deepspeed} for memory-efficient training. In stage 3, we use the LoRA techniques~\cite{hu2022lora} for memory-efficient training. For other stages, we use full fine-tuning. Other training details are shown in~\cref{tab:training-settings}.

\begin{table*}[t]
\centering
\resizebox{\textwidth}{!}{%
\begin{tabular}{|l|c|c|c|}
\hline
\textbf{Setting} & \textbf{Stage 1: Spatial Encoder} & \textbf{Stage 2: Feature Learning} & \textbf{Stage 3: Instruction Finetuning} \\
\hline
\textbf{Hardware} & 8$\times$NVIDIA RTX A6000 (48GB) & 3$\times$NVIDIA A100-SXM (40GB) & 3$\times$NVIDIA A40 (48GB) \\
\textbf{Batch size} & 16 & 12 & 48 \\
\textbf{DeepSpeed} & ZeRO2 & ZeRO2 & ZeRO2 \\
\textbf{LoRA} & - & - & rank=128, alpha=256 \\
\textbf{Epochs} & 10 & 1 & 1 \\
\textbf{Precision} & Full (fp32) & Half (fp16) & Half (fp16) \\
\textbf{Learning rate} & 1e-4 & 2e-5 & 1e-4 (projector: 2e-5) \\
\textbf{Warmup ratio} & 0.1 & 0.03 & 0.03 \\
\textbf{Weight decay} & 0.0 & 0.0 & 0.01 \\
\textbf{Training time} & 1 day & 7.5 days & 19 hours\\
\hline
\end{tabular}}
\caption{Training settings for the three-stage training paradigm.}
\label{tab:training-settings}
\end{table*}

\section{Qualitative analysis on RoI detection}
\label{app:detect-qualitative}


\begin{figure*}[h]
\centering{\includegraphics[width=1.0\textwidth]{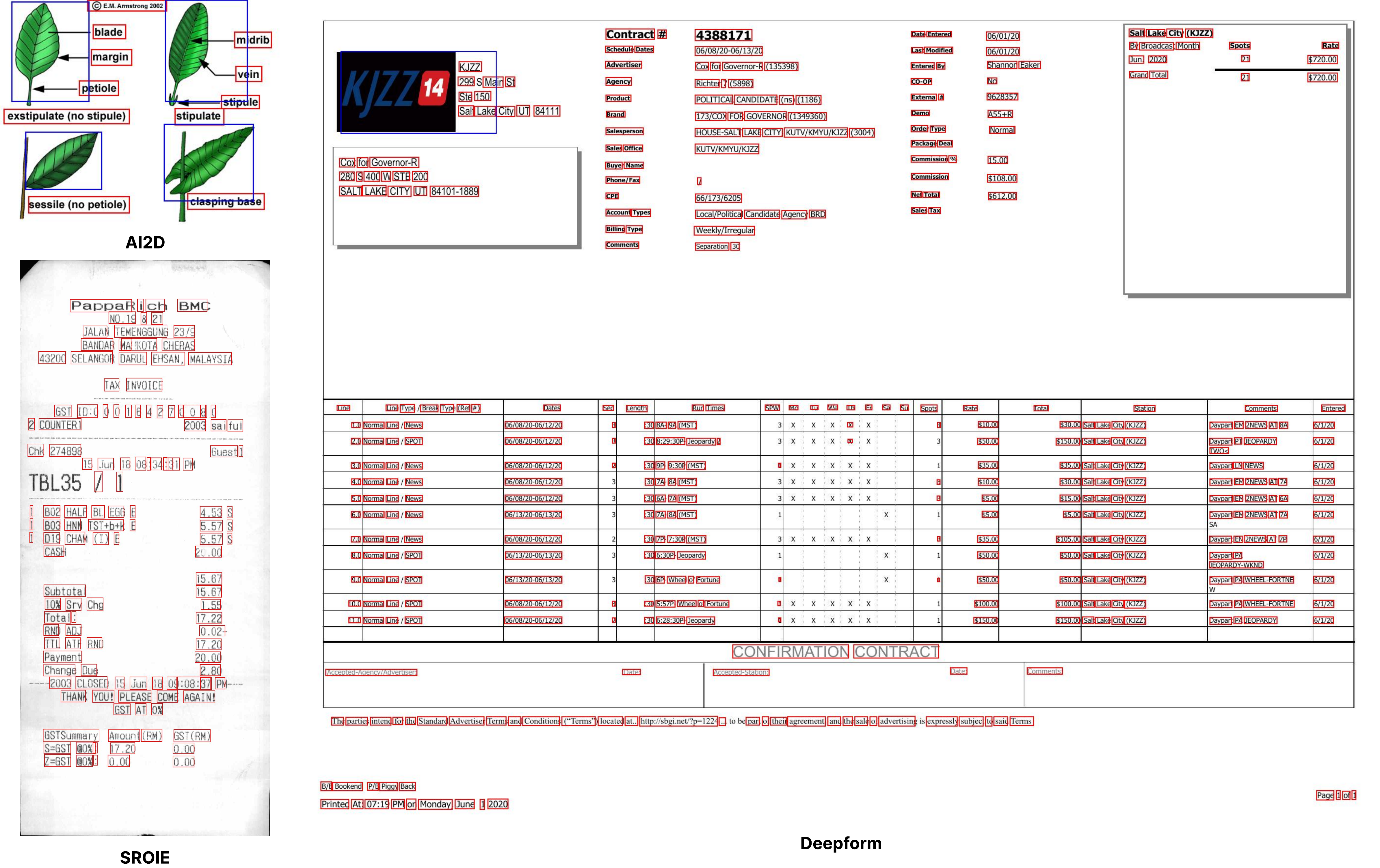}}
\caption{Qualitative results on multimodal RoI detection. The red boxes indicate the text regions, while the blue boxes indicate the vision regions. In general, the RoI detector module performs well in various document types with different resolutions.}
\label{fig:detection_qualitative_results}
\end{figure*}

\cref{fig:detection_qualitative_results} demonstrates the qualitative results of the ROI detector module in the spatial encoder on various document types. Since it is trained on diverse datasets, the ROI detector can detect both word-level text contents and visual objects, such as different types of leaves on the AI2D example. The ROI detector can also process various document types, including receipts in the wild (SROIE sample) or scanned documents (Deepform samples), at different resolutions.

\end{document}